\documentclass[10pt,twocolumn,letterpaper]{article}

\usepackage[pagenumbers]{cvpr} 
\usepackage[table,dvipsnames]{xcolor}
\usepackage{colortbl}
\definecolor{mygrey}{rgb}{0.9,0.9,0.9}
\usepackage{graphicx}
\usepackage{subcaption}
\usepackage{amsmath}
\usepackage{amssymb}
\usepackage{booktabs}
\usepackage{bm}
\usepackage{mathtools}
\usepackage{algpseudocode}
\usepackage{algorithm}
\usepackage{siunitx}
\usepackage{multirow}
\usepackage{enumitem}
\usepackage{caption}
\usepackage{diagbox}
\usepackage[dvipsnames]{xcolor}
\definecolor{myGreen}{RGB}{0,114,0}
\usepackage[pagebackref,breaklinks,colorlinks,citecolor=myGreen]{hyperref}

\usepackage[capitalize]{cleveref}
\crefname{section}{Sec.}{Secs.}
\Crefname{section}{Section}{Sections}
\Crefname{table}{Table}{Tables}
\crefname{table}{Tab.}{Tabs.}

\def\paperID{2812} 
\def\confName{CVPR}
\def\confYear{2025}

\graphicspath{ {images/} }

\algnewcommand\algorithmiclocalize{\textbf{Localize neuron and update model:}}
\algnewcommand\Localize{\item[\algorithmiclocalize]}
\algnewcommand\algorithmiccross{\textbf{Cross-section plane determination:}}
\algnewcommand\Cross{\item[\algorithmiccross]}
\algnewcommand\algorithmicbranchc{\textbrowcolorf{Bifurcation candidates detection:}}
\algnewcommand\Branchc{\item[\algorithmicbranchc]}

\newcommand{\diag}[1]{\text{diag}}
\newcommand{\conj}[1]{\text{conj}}

\usepackage{pifont}
\usepackage{enumitem}
\let\oldding\ding
\renewcommand{\ding}[2][1]{\scalebox{#1}{\oldding{#2}}}

\makeatletter

\newcommand{\prumergepp}[0]{\texttt{freePruner}}
\newcommand{\prumerge}[0]{\texttt{PruMerge}}

\usepackage{soul}

\begin{document}

\title{\prumergepp: \\A Training-free Approach for Large Multimodal Model Acceleration}

\author{
  \textbf{Bingxin Xu$^{1}$}~~~~\textbf{Yuzhang Shang$^{1}$}~~~~\textbf{Yunhao Ge$^{2}$}~~~~\textbf{Qian Lou$^{3}$}~~~~\textbf{Yan Yan$^{1}$}\\
  {\small $^{1}$Illinois Institute of Technology~~~$^{2}$University of Southern California~~~$^{3}$University of Central Florida}\\
\tt\small{victoriaxu@gmail.com}
}

\maketitle

\renewcommand{\thefootnote}{\fnsymbol{footnote}}

\begin{abstract}
Large Multimodal Models (LMMs) have demonstrated impressive capabilities in visual-language tasks but face significant deployment challenges due to their high computational demands. While recent token reduction methods show promise for accelerating LMMs, they typically require extensive retraining or fine-tuning, making them impractical for many state-of-the-art models, especially those with proprietary training data.
We propose \prumergepp, a \textbf{training-free} token reduction approach that can be directly applied to any open-source LMM without additional training. Unlike existing methods that rely heavily on token merging operations, \prumergepp~employs a two-stage token selection strategy: (1) identifying pivotal tokens that capture high-level semantic information using our designed \emph{contribution degree metric}, and (2) selecting complementary tokens that preserve essential low-level visual details through attention pattern analysis. Extensive experiments demonstrate that \prumergepp~achieves \textbf{2× acceleration} while maintaining comparable performance across mainstream visual question-answering benchmarks in the training-free setting. Moreover, \prumergepp~is orthogonal to and can be combined with other post-training acceleration techniques, such as post-training quantization, providing a practical solution for efficient LMM deployment.
\end{abstract}

\section{Introduction}
\label{sec:intro}

\begin{figure}
    \centering
    \vspace{-0.25in}
    \includegraphics[width=0.45\textwidth]{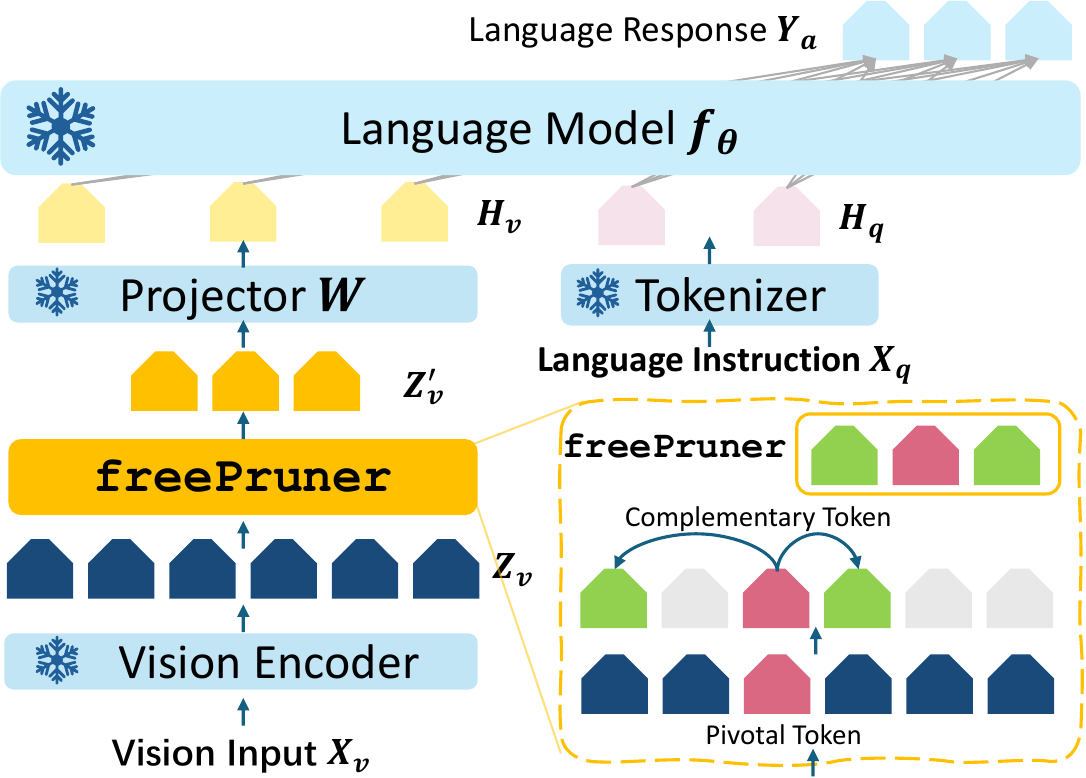}
    \captionsetup{font=small}
        \caption{\textbf{A training-free approach for LMM acceleration.} \colorbox[rgb]{1.0, 0.75, 0.0}{\textbf{\prumergepp}}~enables automatic token selection using only pretrained LMMs, requiring no additional training or fine-tuning. Our method achieves acceleration while maintaining model performance, providing ``free-lunch'' speedup for LMM inference.}
    \vspace{-0.25in}
    \label{fig:intro_brief}
\end{figure}

Large Language Models (LLMs) have demonstrated remarkable capabilities in natural language understanding and generation tasks~\cite{gpt4,team2023gemini,jiang2023mistral,touvron2023llama}.
Building upon these advances, Large Multimodal Models (LMMs) have emerged as powerful tools that integrate visual understanding with language processing~\cite{liu2023llava,liu2024llava,zhu2023minigpt}. By connecting visual encoders with LLMs, these models tackle complex visual tasks such as visual question-answering, image captioning, and visual reasoning~\cite{liu2024llavanext,gpt4,team2023gemini,bai2023qwen}.

Despite their impressive capabilities, the deployment of LMMs faces significant challenges due to their high computational demands. 
This challenge is further amplified in real-world applications where rapid response times and resource efficiency are crucial. 
Several approaches have been proposed to address the efficiency challenges in LMMs as surveyed in \cite{jin2024efficientlmmsurvey}. Methods such as model compression~\cite{frantar2022gptq,lin2024awq,wang2024q,sun2023simple}, and architectural optimization~\cite{chu2023mobilevlm,chu2024mobilevlm} have shown promise. 
\textit{Beyond the obvious efficiency bottleneck in the LLM backbone, LMMs face an additional challenge: the large number of visual tokens.} The computational complexity in the LLM backbone grows quadratically with the number of input tokens, making inference particularly expensive for high-resolution images and videos. Recent works~\cite{shang2024llava,shi2023crossget} have demonstrated that visual tokens, which serve as prefix content for LMMs, often contain substantial redundancy that could be optimized for more efficient processing.

However, existing token reduction approaches typically require extensive retraining or fine-tuning of the LMMs \cite{shang2024llava,shi2023crossget,bolya2023tome,zhang2024token,li2024tokenpacker,liang2022not}, which presents significant practical limitations. Specifically, reducing the number of multimodal tokens necessitates retraining the entire model on the original dataset to adapt to the reduced token setting. \textbf{This retraining requirement poses substantial practical challenges.} State-of-the-art LMMs undergo extensive visual instruction fine-tuning using large-scale datasets, requiring intensive computational resources. For instance, training LLaVA-OneVision-72B~\cite{li2024llava} demands thousands of A100 GPU hours, while even the more modest LLaVA-1.5-13B~\citep{liu2024llava} requires hundreds of hours on A100 GPUs.
Unfortunately, this represents the best-case scenario, where training materials—data and training recipes—are publicly accessible. More critically, many high-performing LMMs rely on proprietary training data that is not publicly available, making reproduction or retraining impractical for the broader research community. For example, powerful models like Pali-Gemma~\cite{beyer2024paligemma}, Intern VL2~\cite{chen2024internvl2}, and Qwen VL2~\cite{wang2024qwen2} only release their model weights while keeping their training data private, rendering retraining impossible.

\begin{figure}
    \centering
    \vspace{-0.25in}
    \includegraphics[width=0.4\textwidth]{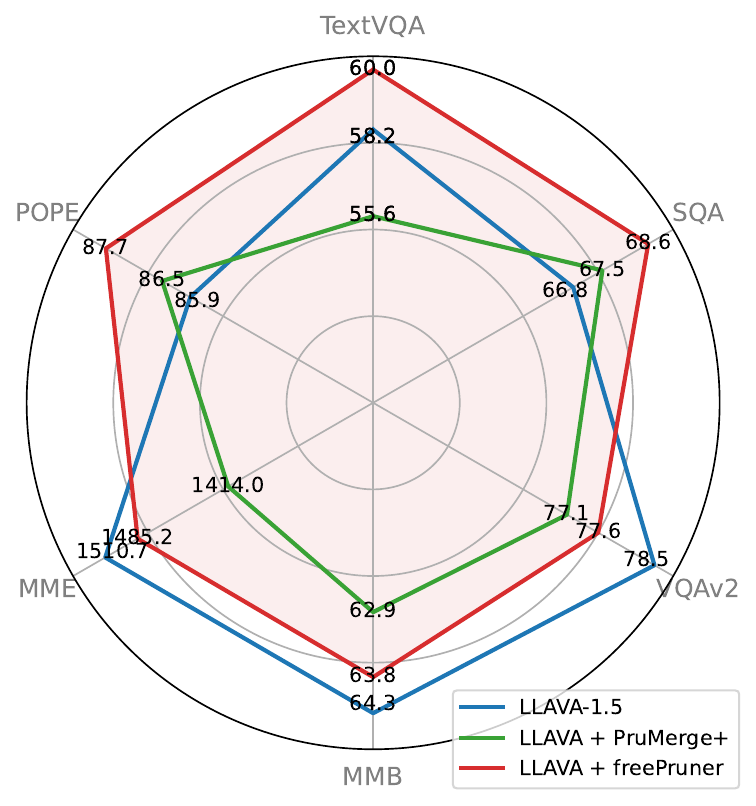}
    \vspace{-0.1in}
    \captionsetup{font=small}
        \caption{\textbf{Performance comparison across VQA tasks.} \prumergepp~achieves 2× acceleration of LLaVA while maintaining comparable performance across mainstream VQA benchmarks. Notably, our token selection approach is orthogonal to other training-free LLM acceleration methods (e.g., quantization), enabling potential combinations for even greater efficiency gains.
        }
    \vspace{-0.2in}
    \label{fig:intro_results}
\end{figure}

To address these challenges, we propose \prumergepp, a training-free approach that can be directly applied to any open-source LMM without additional training or fine-tuning. Through extensive analysis of previous token reduction studies~\cite{shang2024llava,shi2023crossget,bolya2023tome,zhang2024token,li2024tokenpacker,liang2022not}, we identify a critical limitation: existing methods rely heavily on token merging operations. While these operations effectively pack the information into compressed tokens, they substantially change the distribution of token representation. This alteration makes it challenging for pretrained models to handle the modified representations without additional retraining. This observation explains why previous token reduction methods perform poorly in training-free settings and require full model retraining.
In contrast, \prumergepp~ employs a pure token \emph{selection} strategy, eschewing any merging operations.
To ensure comprehensive visual understanding, our approach captures both high-level and low-level representations through two key components. First is \textbf{Pivotal Token Selection}. We design a token contribution degree metric to identify tokens that capture high-level semantic information across multiple transformer layers \cite{kim2024vision,cai2022coarse}.
Next is \textbf{Complementary Token Selection}. Based on attention patterns in the penultimate layer, we select additional tokens that exhibit strong relationships with pivotal tokens, preserving essential low-level visual details.

Our main contributions are threefold:
\begin{itemize}
    \item We propose a training-free paradigm for accelerating LMMs that applies to any open-sourced LMMs, requiring no access to training data or further fine-tuning.
    \item We propose a two-stage token selection strategy that balances high-level semantic features with low-level visual details. This approach achieves approximately 2× acceleration while maintaining performance (see Fig.~\ref{fig:intro_results}).
    \item \prumergepp~is orthogonal to existing post-training acceleration methods such as quantization~\cite{lin2024awq,shang2023pb,shang2024enhancing}, offering an additional option for LMM acceleration.
\end{itemize}

\section{Related Work}
\label{sec:related}

\noindent\textbf{Token Reduction for Efficient LMMs}

Large Language Models (LLMs) such as GPT-4~\citep{gpt4}, LLaMA~\citep{touvron2023llama}, Mistral~\citep{jiang2023mistral}, and Gemini~\citep{team2023gemini} have demonstrated remarkable capabilities in text-based question answering and reasoning tasks. Building upon these advances, Large Multimodal Models (LMMs)~\citep{liu2023llava,zhu2023minigpt,yin2023survey,zhang2024mm} extend these capabilities to visual domains, enabling chat-based interactions that combine image understanding with language generation. Recent developments have further expanded LMM capabilities to include region-level understanding~\citep{cai2023vipllava,zhang2023gpt4roi,peng2023kosmos,chen2023shikra}, video comprehension~\citep{lin2023video,zhang2023video}, and 3D scene interpretation~\citep{3dllm}.
These multimodal architectures typically process visual information by feeding visual tokens directly into the LLM as prefix tokens, utilizing various interface mechanisms such as MLPs~\citep{liu2023improvedllava}, Qformers~\citep{instructblip, zhu2023minigpt}, or resamplers~\citep{alayrac2022flamingo}. However, the number of visual tokens can become prohibitively large, particularly for high-resolution images~\citep{liu2024llava,GPT4V_System_Card}. This poses a significant challenge due to the quadratic computational complexity of Transformer architectures~\citep{vaswani2017attention}.

Token reduction has emerged as a promising approach for accelerating Transformer-based models~\citep{haurum2023tokens}. Traditional uni-modal token reduction methods~\citep{liu2022adaptive,yin2022vit,liang2022not,bolya2023tome,fayyaz2022adaptive} focus on reducing token quantity within the internal transformer structure to decrease computational overhead. For example, token merging~\citep{bolya2023tome} maintains full attention while progressively reducing tokens through a bipartite matching-based selection of representative tokens.
In the context of LMMs, several recent approaches have been proposed for token reduction. PruMerge~\cite{shang2024llava} and TokenPacker~\cite{li2024tokenpacker} achieve compression rates of 20\% to 50\% through token aggregation based on similarity metrics. Qwen-VL~\cite{bai2023qwen} employs a resampler to compress visual tokens to a fixed length. However, these methods face significant limitations in training-free settings due to their reliance on token merging operations. While merging effectively condenses information, it fundamentally alters the token distribution, requiring model retraining to maintain performance. This limitation makes existing approaches impractical for scenarios where retraining is not feasible.
To the best of our knowledge, our work presents the first exploration of training-free token reduction specifically designed for LMMs.
\section{\prumergepp: A Training-free Approach for Large Multimodal Model Acceleration}
\label{sec:method}

\begin{figure}[!t]
\centering
  \includegraphics[width=0.48\textwidth]{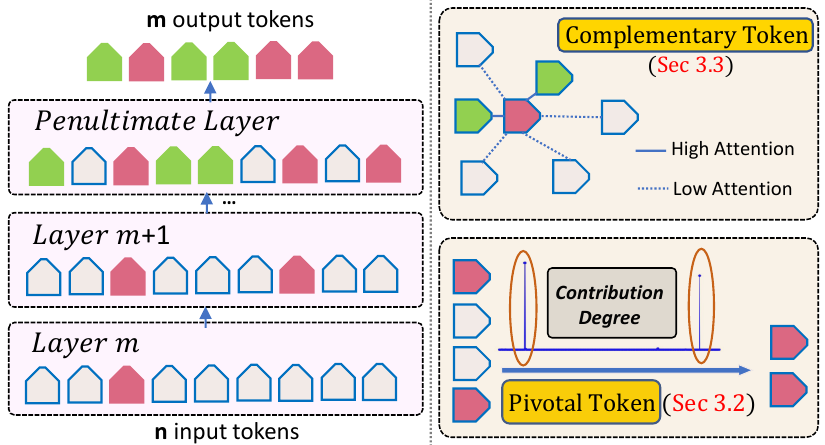}
  \caption{\textbf{The overview of \prumergepp.} \prumergepp~has 2 modules: (1)Identify \colorbox[rgb]{0.8588, 0.4078, 0.5098}{pivotal tokens} for high-level visual information via  contribution degree across layers (see Sec.\ref{subsec:pt}). Our designed token contribution degree matrics distributes sparsely in the encoder's middle layers, thus we leverage this property to select the tokens representing high-level features; (2)Select complementary tokens for low-level visual information in the penultimate layer (see Sec.\ref{subsec:ct}). Via this module, we can further use the pivotal tokens as anchors to retrieve \colorbox[rgb]{0.5725, 0.8157, 0.3137}{complementary tokens} containing low-level feature information. In this way, we can realize the training-free token selection in a coarse-to-fine manner.}
  \vspace{-0.15in}
 \label{fig:pipeline}
\end{figure}

In this paper, we propose a post-training LMM acceleration method in a training-free environment.
In this section, we first revisit the overall structure of Large Multimodal Models (LMMs) and the architecture of transformer encoders. We emphasize the direct relationship between the number of visual tokens and the computational efficiency of LMMs (Sec.\ref{subsec:prelim}). 
Subsequently, we present our training-free token reduction method, \prumergepp, designed specifically for LMM acceleration. 
Our method features two key components: 
(1) Pivotal token identification, which extracts tokens containing coarse-to-fine feature information from the visual encoders (Sec.\ref{subsec:pt}); (2) Complementary token selection, which utilizes pivotal tokens as anchors to select highly correlated visual tokens (Sec.\ref{subsec:ct}). The pipeline is visualized in Fig.\ref{fig:pipeline}.

\subsection{Preliminaries}
\label{subsec:prelim}

\textbf{Large Multimodal Models (LMMs).} The fundamental concept behind LMMs is to leverage large language models (LLMs) for multimodal understanding, as illustrated in Fig.~\ref{fig:intro_brief} (without \prumergepp~part). 
Given a multimodal input $\mathbf{X}_{v}$ (e.g., an image or video), the system first employs a Transformer encoder~\cite{vaswani2017attention,dosovitskiy2020image} to generate multimodal tokens $\mathbf{Z}_{v}$. These tokens are then processed through a projector $\mathbf{W}$ that maps them into the text feature space. After aligning the visual tokens $\mathbf{H}_{v}$ and textual tokens $\mathbf{H}_{q}$ in the same feature space, they are concatenated and processed by the LLM backbone \( f_{\theta} \) to generate the final outputs  $\mathbf{Y}_{a}$~\citep{zhang2024mm}.

\noindent\textbf{Transformer Encoder}~\citep{dosovitskiy2020image,vaswani2017attention} serve as the bridge to transform a visual input to visual tokens representation, which later is sent to a LLM for understanding~\citep{liu2024llava,zhu2023minigpt,3dllm,zhang2024mm,lin2023video}. 
It is commonly used as visual encoders' architecture in LMMs. It is mainly composed of multiple transformer blocks that include multi-head self-attention (MSA) layers, a feed-forward neural network (FFN), skip connections, and layer normalization ~\citep{ba2016layer,vaswani2017attention}. In the encoder, an visual input is first segmented into a grid of patches, each of which is then transformed into token embeddings. As these tokens pass through successive transformer layers, their representations become progressively refined. Within the self-attention layer, each input token is transformed into three distinct vectors: query $\mathbf{q}$, key $\mathbf{k}$, and value $\mathbf{v}$, via corresponding linear transformation matrices $\mathbf{W}_q$, $\mathbf{W}_k$, and $\mathbf{W}_v$. 
Applying these transformation vectors on the inputs leads to the matrices for query, key and value: $\mathbf{Q}$, $\mathbf{K}$, and $\mathbf{V}$. The attention matrix $\mathbf{A}$ measures how much attention it contributes to other elements:
\begin{equation}
    \mathbf{A} = \text{softmax}\left(\frac{\mathbf{Q} \cdot \mathbf{K}^\mathbf{T}}{\sqrt{d_k}}\right)
    \label{eq:attentionmatrix}
\end{equation}
where $d_k$ is the dimension of $\mathbf{q}$ and $\mathbf{k}$. A higher value in $\mathbf{A}$ indicates a greater relevance between two tokens.
Self-attention generates new representations of each token by aggregating information from all other tokens, weighted by the attention scores:
\begin{equation}
    \mathbf{Y} = \mathbf{A} \cdot \mathbf{V}
    \label{eq:selfattention}
\end{equation}
Following the self-attention layers, the transformer encoder incorporates a feed-forward network (FFN). This network comprises two layers of linear transformations that are interspaced by a nonlinear activation function, expressed as:
$\text{FFN}(\mathbf{X}) = \mathbf{W}_2 \sigma(\mathbf{W}_1 \mathbf{X})$.
where $\mathbf{W}_1$ and $\mathbf{W}_2$ are the matrices of the linear transformation layers, and $\sigma$ denotes the nonlinear activation function. 

A critical challenge in the LMM architecture is that the computational complexity grows quadratically with the number of input tokens~\citep{tay2022efficient}. Since visual tokens comprise the majority of these inputs for LLM backbone, reducing visual token quantity is crucial for improving LMM efficiency~\cite{shang2024llava,shi2023crossget}. \textbf{However, existing token reduction approaches~\cite{shang2024llava,shi2023crossget,bolya2023tome,li2024tokenpacker,zhang2024token} are not suitable for training-free LLM acceleration scenarios.}
This limitation stems from their reliance on token merging operations, which fundamentally alter the token distribution ($p_{\mathbf{Z}_{v}}$).
Specifically, these methods typically combine multiple tokens through weighted summation or more complex operations to create compressed representations. 
While such merging operations can theoretically preserve information in a more condensed form, they introduce perturbations to the original token distribution.
These perturbations cause \textit{a gap between the merged tokens and the expected input distribution of the pretrained projector $\mathbf{W}$ and LLM \( f_{\theta} \)}.
Consequently, the merged representations deviate significantly from the distribution the model was trained on, making them difficult to process without model retraining.

To achieve training-free token reduction while preserving essential information for the LLM backbone, we must first examine the internal architecture of multimodal encoders, specifically the Transformer architecture~\citep{dosovitskiy2020image}.

\subsection{Pivotal Token for High-level Feature}
\label{subsec:pt}

A crucial aspect of LMM's multimodal understanding is the effective balance between low-level and high-level visual features~\cite{cai2023vipllava,wang2025sclip}. While previous works~\cite{lioubashevski2024looking,zhou2022maskclip} have explored various approaches to feature extraction, \textit{maintaining this balance in a training-free token selection setting presents unique challenges}. 
The primary challenge lies in identifying and selecting the most informative tokens using only the internal representations and attention patterns within the transformer, without access to additional training signals or external supervision. 
To address this, we first introduce a metric called the \textbf{token contribution degree} $\mathbf{r}_l$, which quantifies how much each token influences other tokens in the network. This metric is defined as:
\begin{equation}
    \mathbf{r}_l^i = \sum \mathbf{A}_{l}[:,i] - \mathbf{A}_{l}[i,i],
    \label{eq:contribution_degree}
\end{equation}
where $\mathbf{A}_{l}$ represents the attention map at layer $l$, computed using Eq.\ref{eq:attentionmatrix}. This metric measures the extent to which the $i$-th token contributes to the entire image representation at layer $l$. 
The intuition behind this metric is straightforward: a larger $\mathbf{r}_l^{i}$ indicates that in layer $l$, most tokens direct their attention to the $i$-th token, suggesting that $i$-th token effectively represents the features extracted at that layer. 

\begin{figure}[!t]
    \centering
    \begin{subfigure}[b]{0.48\textwidth}
        \centering
        \includegraphics[width=0.98\textwidth]{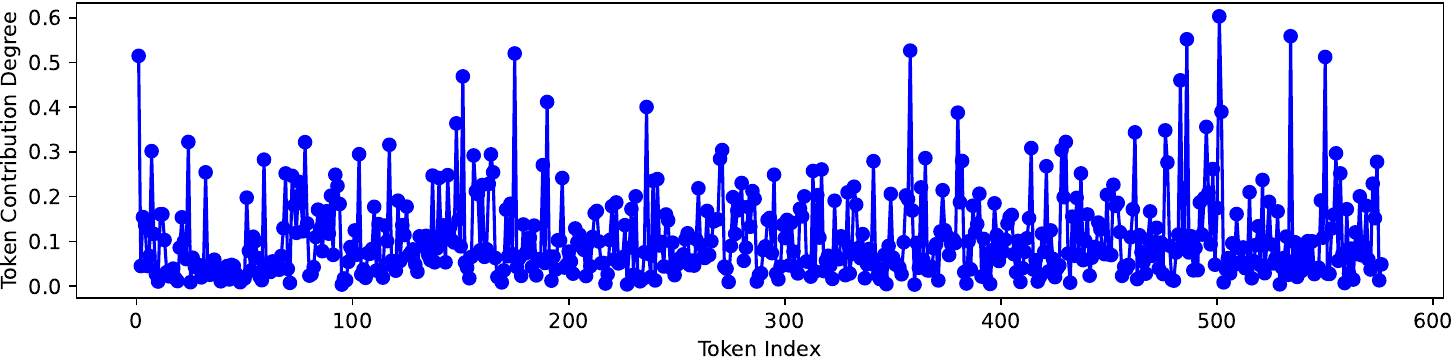}
        \vspace{-0.1cm}
        \subcaption{The distribution of token contribution degree in 6-th layer.}
        \vspace{0.1cm}
        \label{subfig:1}
    \end{subfigure}
    \hfill
    \begin{subfigure}[b]{0.48\textwidth}
        \centering
        \includegraphics[width=0.98\textwidth]{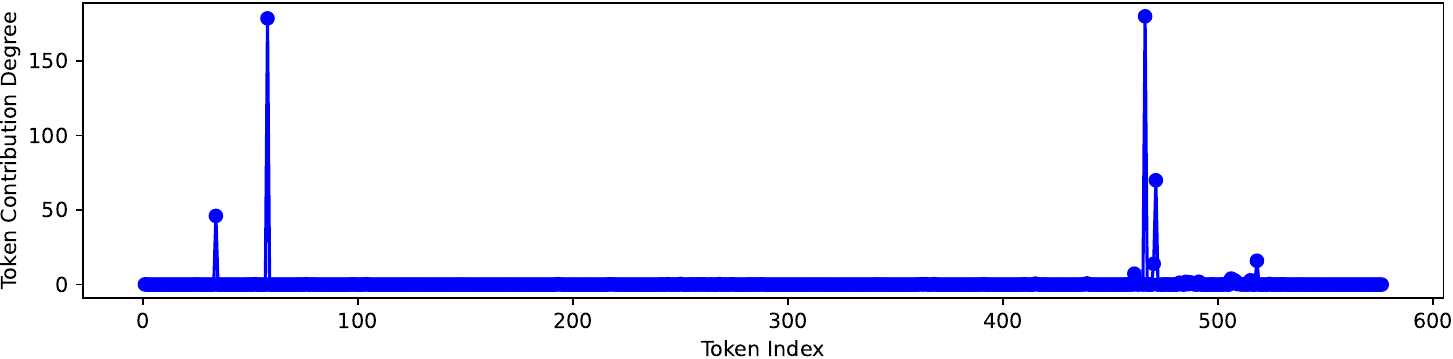}
        \vspace{-0.1cm}
        \subcaption{The distribution of token contribution degree in 12-th layer.}
        \vspace{0.1cm}
        \label{subfig:2}
    \end{subfigure}
    \hfill
    \begin{subfigure}[b]{0.48\textwidth}
        \centering
        \vspace{-0.1cm}
        \includegraphics[width=0.98\textwidth]{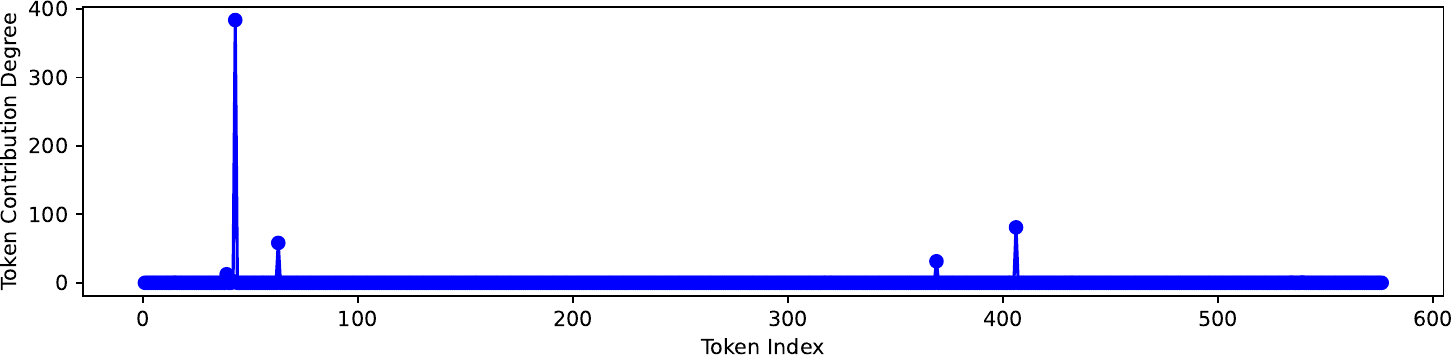}
        \subcaption{The distribution of token contribution degree in 18-th layer.}
        \vspace{0.1cm}
        \label{subfig:3}
    \end{subfigure}
    \hfill
    \begin{subfigure}[b]{0.48\textwidth}
        \centering
        \vspace{-0.1cm}
        \includegraphics[width=0.98\textwidth]{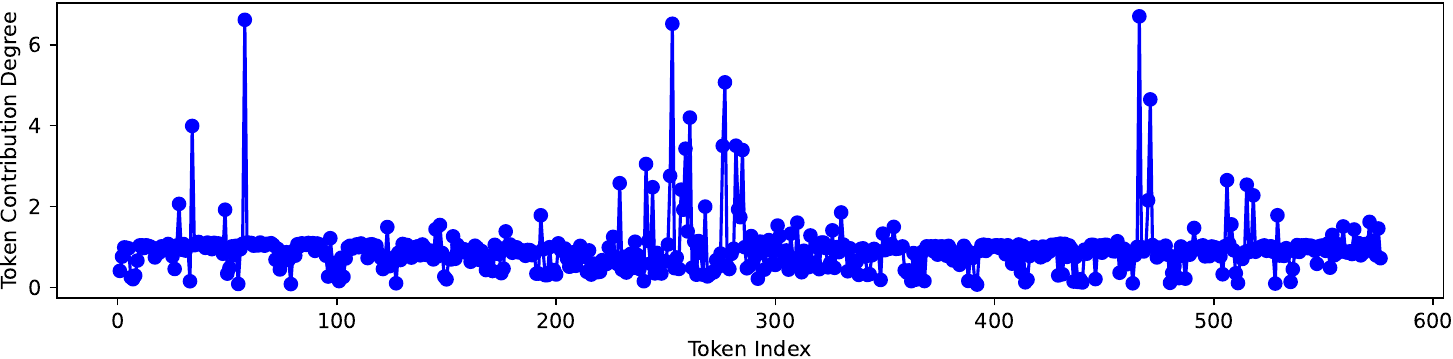}
        \subcaption{The distribution of token contribution degree in 23-th layer.}
        \vspace{0.1cm}
        \label{subfig:4}
    \end{subfigure}
    \caption{Distribution of token contribution degree $\mathbf{r}_l$ across different transformer layers. The consistently sparse distribution patterns demonstrate that only a small subset of tokens serves as global information aggregators at each layer, regardless of network depth. This sparsity property enables effective identification of pivotal tokens for high-level feature representation.}
    \vspace{-0.2in}
\label{fig:attn_dirtibution}
\end{figure}

Fig.\ref{fig:attn_dirtibution} presents the distribution of token contribution degrees across different layers. Notably, in most layers, $\mathbf{r}_l$ exhibits a highly sparse distribution, with only a few tokens showing significant contribution degrees. This sparsity pattern is consistent across different depths of the network, from shallow (layer 6) to deep layers (layer 23). The observed sparsity reveals an important property: in the middle layers, while most visual tokens primarily attend to themselves, a small subset of pivotal tokens emerges that significantly influences the representations of other tokens.

Fig.\ref{fig:attn_dirtibution} also illustrates varying degrees of contribution across different layers. In the shallower layers, the contribution degree is minimal for all tokens, suggesting that no single token significantly influences others. As the layers deepen, the contribution degree noticeably increases, suggesting a progressive sharing of information among tokens. A contribution level of 400 means that, on average, a pivotal token delivers an attention score of 0.7 to each visual token, thus contributing to 70\% of the information exchanged among all tokens. However, the contribution level diminishes in the penultimate layer. At this stage, having interacted through all previous layers, each token has garnered extensive information and has developed a distinct representation, leading to a more balanced interaction where no single token predominates.

Building upon this observation, we propose a selective token retention strategy based on the token contribution degree $\mathbf{r}_l$. As shown in Fig.\ref{fig:pipeline}, our approach strategically identifies and retains tokens with high contribution degrees from deeper layers to capture high-level semantic features essential for complex visual understanding. 

\noindent\textbf{Intuitive Explanation:} These high-contribution visual tokens function analogously to global tokens within the visual token set, similar to the class token in vision transformer architectures. Just as the class token in vision transformers~\cite{dosovitskiy2020image} aggregates information from all tokens for classification, our identified high-contribution tokens serve as natural focal points that integrate information from surrounding tokens. This emergent behavior allows these tokens to capture global context effectively, making them particularly valuable for representing complex visual features across different hierarchical levels. 

\begin{figure}[!t]
    \centering
    \includegraphics[width=0.48\textwidth]{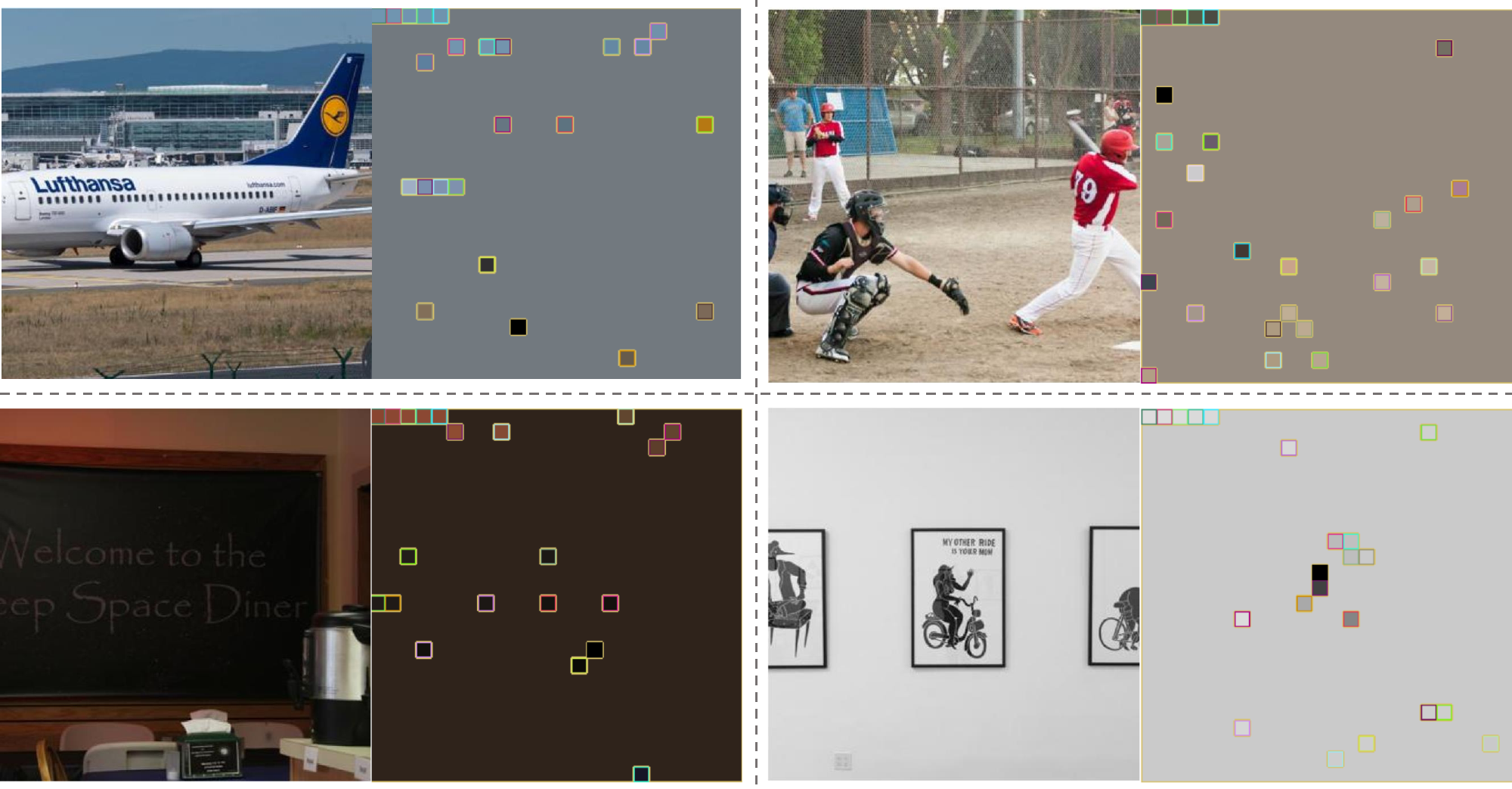}
    \caption{Pivotal tokens are found in areas of the image with dense information. They capture the high-level semantic information.}
    \vspace{-0.2in}
\label{fig:pt_visualziation}
\end{figure}

We present the positioning of pivotal tokens in Fig.\ref{fig:pt_visualziation}. Most pivotal tokens are located in the image areas that contain the most information. However, there is one counter-intuitive observation. Initial tokens, typically positioned in the top left corner with minimal visual content such as a plain white-wall background, are often marked as pivotal. This observation aligns with insights from existing studies on Transformers. For example, retaining initial tokens is noted to significantly enhance window attention (\ie, attention sink), despite their lack of semantic importance~\cite{xiao2023efficient}. This supports our approach to selecting initial tokens as pivotal, emphasizing their high contribution degree and their role in improving performance in subsequent tasks.




\begin{figure}[!t]
    \centering
    \begin{subfigure}[b]{0.48\textwidth}
        \centering
        \includegraphics[width=0.98\textwidth]{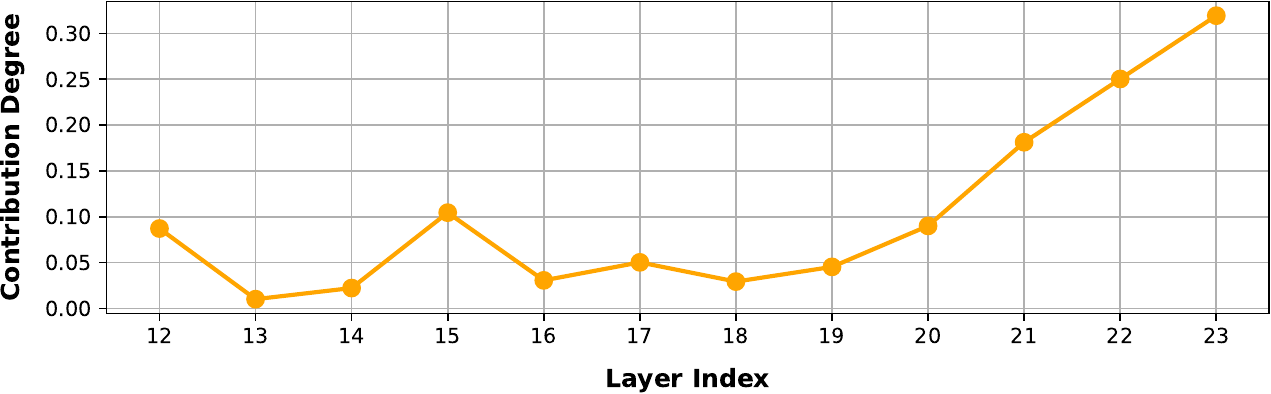}
        \subcaption{Complementary token's contribution degree $\mathbf{r}_l$ across layers.}
        \label{subfig:5}
    \end{subfigure}
    \hfill
    \begin{subfigure}[b]{0.48\textwidth}
        \centering
        \includegraphics[width=0.98\textwidth]{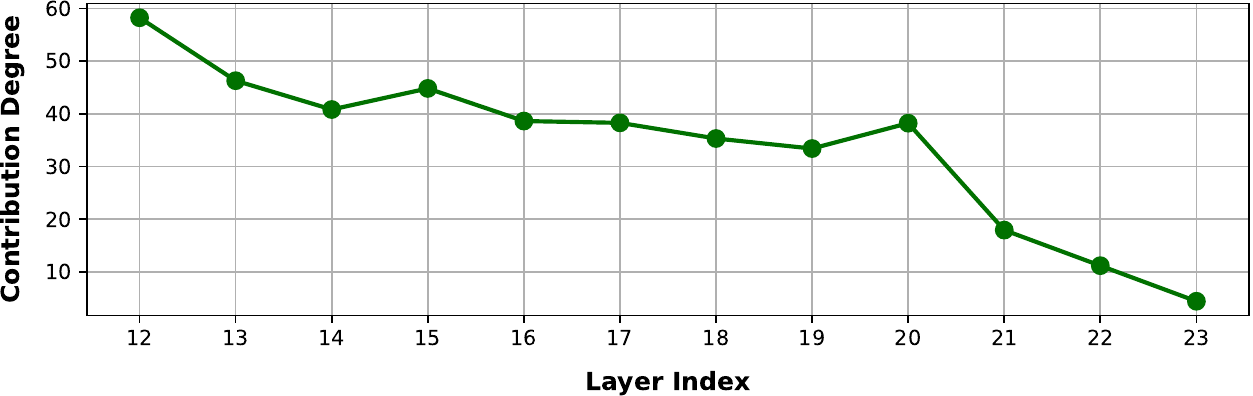}
        \subcaption{Pivotal token's contribution degree $\mathbf{r}_l$ across layers.}
        \label{subfig:6}
    \end{subfigure}
    \caption{Token contribution degree $\mathbf{r}_l$ exhibits divergent trends across transformer layers for pivotal and complementary tokens. Pivotal tokens demonstrate a decreasing trend in contribution degree as network depth increases, whereas complementary tokens increasingly contribute more in deeper layers.}
    \vspace{-0.1in}
\label{fig:ct_pt}
\end{figure}

\begin{algorithm}[!t]  
\caption{\prumergepp~algorithms for reducing the number of visual tokens.}  
\label{alg:pru-merging}
\begin{algorithmic}[1] 
	\Require The original representative tokens, $\mathbf{Y}=\{\mathbf{y}_1,\cdots \mathbf{y}_n\}$. $n$ is the number of input visual tokens. 
    $L$ is the total number of layers, and $L_s$ is the index of the starting layer for pivotal token identification.
	\Ensure Refine $\mathbf{Y}$ to $m$ (adaptive) visual tokens $\mathbf{Y}^{\prime}=\{\mathbf{y}_1^{\prime},\cdots \mathbf{y}_m^{\prime}\}$, in which $m < n$.
    \State \textbf{\prumergepp}:
    \For{$l$ in range($L_s$, $L$)}
        \State Calculate token contribution degree $r^{i}_{l}$ for every visual token using Eq.\ref{eq:contribution_degree}.
        \State Select $k$ pivotal token indices based on the highest $r^{i}_{l}$, and obtain pivotal tokens index list $\mathbf{p}_l = \{i_{l,1}, \ldots, i_{l,k}\}$. (see Sec.\ref{subsec:pt})
    \EndFor
    \State The complete indices list of pivotal tokens are $\mathbf{p} = \bigcup_{l \in \left[L_s, L\right]} \mathbf{p}_l$
    \For{$p$ in $\mathbf{p}$}
        \State Adaptively select outliers from the attention score vector $\mathbf{A}[i_p,:]$ of the $i_L$-th layer (obtaining $\mathbf{A}$ with Eq.\ref{eq:attentionmatrix}). Obtain complementary token index list: $\mathbf{c}_l = \{i_{p,1}, \ldots, i_{p,j}\}$ 
    \EndFor
    \State The complete indices list of complementary tokens are $ \mathbf{c} = \bigcup_{l} \mathbf{c}_l$ (see Sec.~\ref{subsec:ct}).
    \State Final indices list of selected tokens $\mathbf{s} = \mathbf{p} \cup \mathbf{c}$.
    \State Output a refined stack of visual tokens $\mathbf{Y}^{\prime}=\{\mathbf{y}_1^{\prime},\cdots \mathbf{y}_m^{\prime}\}$ based on $\mathbf{s}$.
\end{algorithmic}  
\end{algorithm}

\begin{table*}[!th]
\centering
\caption{In a comparative analysis across six benchmarks with large multimodal models, our \prumergepp~ method uses only 50\%  of the input visual tokens on average for these tasks, but delivers competitive performance relative to the original LLaVA-1.5. Note that LLaVA-PruMerge+~\citep{shang2024llava} includes model fine-tuning after token reduction, and LLaVA+PruMerge+ applies token reduction without subsequent fine-tuning. Our method is totally training-free.} 
\scalebox{0.95}{
\begin{tabular}{lllcc|ccc|ccc}
\toprule
Method & LLM & Res. & PT & IT & VQA\textsuperscript{v2} & SQA\textsuperscript{I} & VQA\textsuperscript{T} & POPE & MME & MMB \\
\midrule
BLIP-2 & Vicuna-13B & 224 & 129M & - & 41.0 & 61 & 42.5 & 85.3 & 1293.8 & - \\
InstructBLIP & Vicuna-7B & 224 & 129M & 1.2M & - & 60.5 & 50.1 & - & - & 36 \\
InstructBLIP& Vicuna-13B & 224 & 129M & 1.2M & - & 63.1 & 50.7 & 78.9 & 1212.8 & - \\
Shikra & Vicuna-13B & 224 & 600K & 5.5M & 77.4 & - & - & - & - & 58.8 \\
IDEFICS-9B & LLaMA-7B & 224 & 353M & 1M & 50.9 & - & 25.9 & - & - & 48.2 \\
IDEFICS-80B & LLaMA-65B & 224 & 353M & 1M & 60.0 & - & 30.9 & - & - & 54.5 \\
Qwen-VL & Qwen-7B & 448 & 1.4B & 50M & 78.8 & 67.1 & 63.8 & - & - & 38.2 \\
Qwen-VL-Chat & Qwen-7B & 448 & 1.4B & 50M & 78.2 & 68.2 & 61.5 & - & 1487.5 & 60.6 \\
\hline
LLaVA-1.5 & Vicuna-7B & 336 & 558K & 665K & 78.5 & 66.8 & 58.2 & 85.9 & 1510.7 & 64.3 \\
LLaVA-PruMerge+ & Vicuna-7B & 336 & 558K & 665K & 76.8 & 68.3 & 57.1 & 84.0 & 1462.4 & 64.9 \\
LLaVA + PruMerge+ & Vicuna-7B & 336 & 0 & 0 & 76.6 & 67.5 & 55.6 & 86.5 & 1414.0 & 62.9 \\
\rowcolor{brown!20} LLaVA + \prumergepp & Vicuna-7B & 336 & 0 & 0 & 77.6 & 68.6 & 60.0 & 87.7 & 1485.2 & 63.8 \\
\hline
LLaVA-1.5 & Vicuna-13B & 336 & 558K & 665K  & 80.0 & 71.6 & 61.3 & 85.9 & 1531.3 & 67.7 \\
LLaVA-PruMerge+ & Vicuna-13B & 336 & 558K & 665K  & 77.8 & 71.0 & 58.6 & 84.4 & 1485.5 & 65.7 \\
LLaVA + PruMerge+ & Vicuna-13B & 336 & 0 & 0 & 77.6 & 71.9 & 56.3 & 85.5 & 1470.1 & 64.5 \\
\rowcolor{brown!20} LLaVA + \prumergepp & Vicuna-13B & 336 & 0 & 0 & 78.7 & 72.5 & 60.0 & 87.9 & 1516.7 & 68.0\\
\bottomrule
\end{tabular}}
\label{tab:main_table}
\end{table*}

\subsection{Complementary Token for Low-level Feature}
\label{subsec:ct}
Pivotal tokens distill essential semantic content from visual inputs, focusing primarily on key object features. However, they often miss finer, low-level details, particularly in information-rich images. To address this, we propose a \textbf{complementary token selection} method to target and incorporate low-level features into the image representation. 

As tokens progress through the network layers, they strive to encapsulate key information to render a comprehensive understanding. However, as depicted in Fig.\ref{fig:attn_dirtibution}, certain tokens continue to exhibit high contribution degree in the penultimate layer, indicating that some information remains unabsorbed by the pivotal tokens. This unabsorbed information primarily consists of low-level details that pivotal tokens, with their focus on high-level semantic content, fail to capture. Thus, there is a clear need to integrate the missing low-level information into image representation.

Fig.\ref{fig:ct_pt} demonstrates how the contribution degree vary across network layers, highlighting a completely opposite trend between complementary tokens and pivotal tokens. It reveals that as the network layers deepen, the contribution degree of complementary tokens steadily increases, indicating a consistent release of low-level information. In contrast, pivotal tokens exhibit a high contribution degree in the middle layers, which gradually declines in deeper layers. This shift suggests that the network, initially focused on high-level semantics, gradually incorporates more detailed low-level features, such as textures and edges, which are critical for a refined understanding.

Our selection method for complementary tokens leverages their high attention scores towards pivotal tokens in the penultimate layer, identifying those that carry vital low-level details not captured by pivotal tokens. This approach moves beyond using uniformly sampled spatial tokens, which merely capture spatial data at regular intervals, to a strategy that targets the essential low-level visual details. Complementary tokens are not just served as expanding pivotal tokens but are important in delivering the nuanced low-level information essential for image understanding. Related experiments and discussions are in Sec.\ref{subsec: Ablation Study}.
The complete token selection algorithm is outlined in Alg.\ref{alg:pru-merging}.

\section{Experiments}
We first present the empirical results of our token reduction method, \prumergepp, applied to LLaVA-1.5 in Sec.\ref{subsec: Main Results}. Second, we explore the scalability in Sec.\ref{subsec: Scalability Analysis} and generalization capability in Sec.\ref{subsec:generalization} through various experiments.
Furthermore, we evaluate the efficiency enhancements achieved by employing our \prumergepp~on LMM in Sec.\ref{subsec: efficiency analysis}. Finally, we demonstrate the necessity and effectiveness of each module in our model in Sec.\ref{subsec: Ablation Study}.

\begin{figure*}[!t]
\begin{minipage}{\textwidth}
    \begin{subfigure}{0.33\textwidth}
    \includegraphics[width=\textwidth]{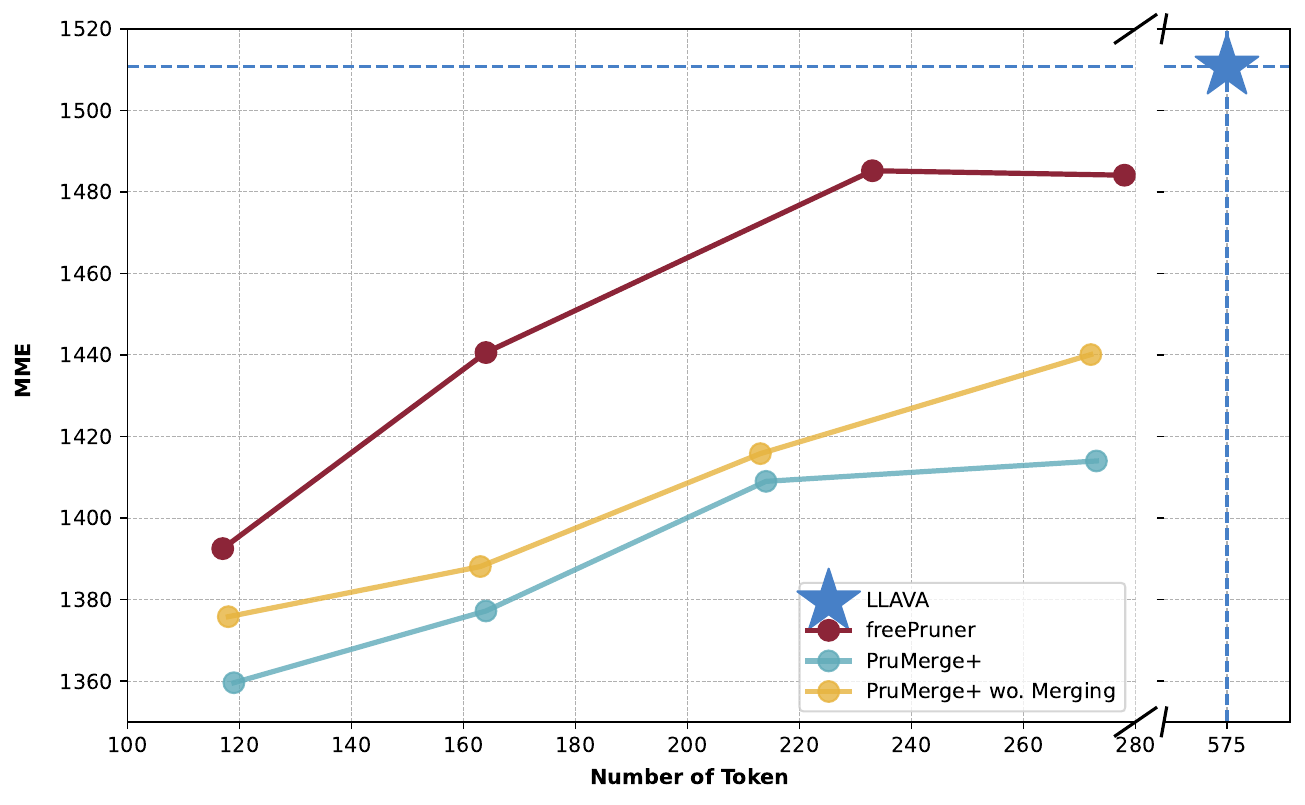}
    \caption{MME~\citep{fu2023mme}}
    \label{subfig1}
    \end{subfigure}
    \hfill
    \begin{subfigure}{0.33\textwidth}
    \includegraphics[width=\textwidth]{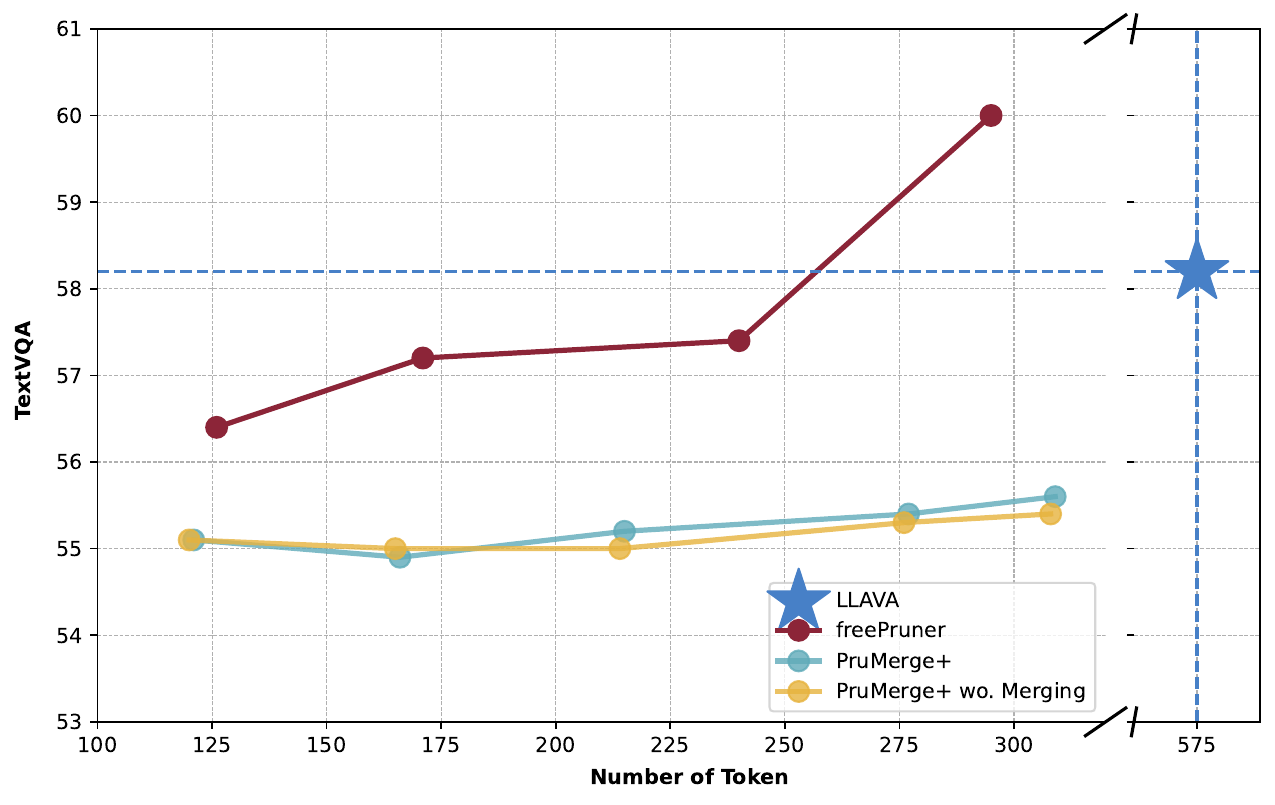}
    \caption{TextVQA~\citep{singh2019textvqa}}
    \label{subfig2}
    \end{subfigure}
    \hfill
    \begin{subfigure}{0.33\textwidth}
    \includegraphics[width=\textwidth]{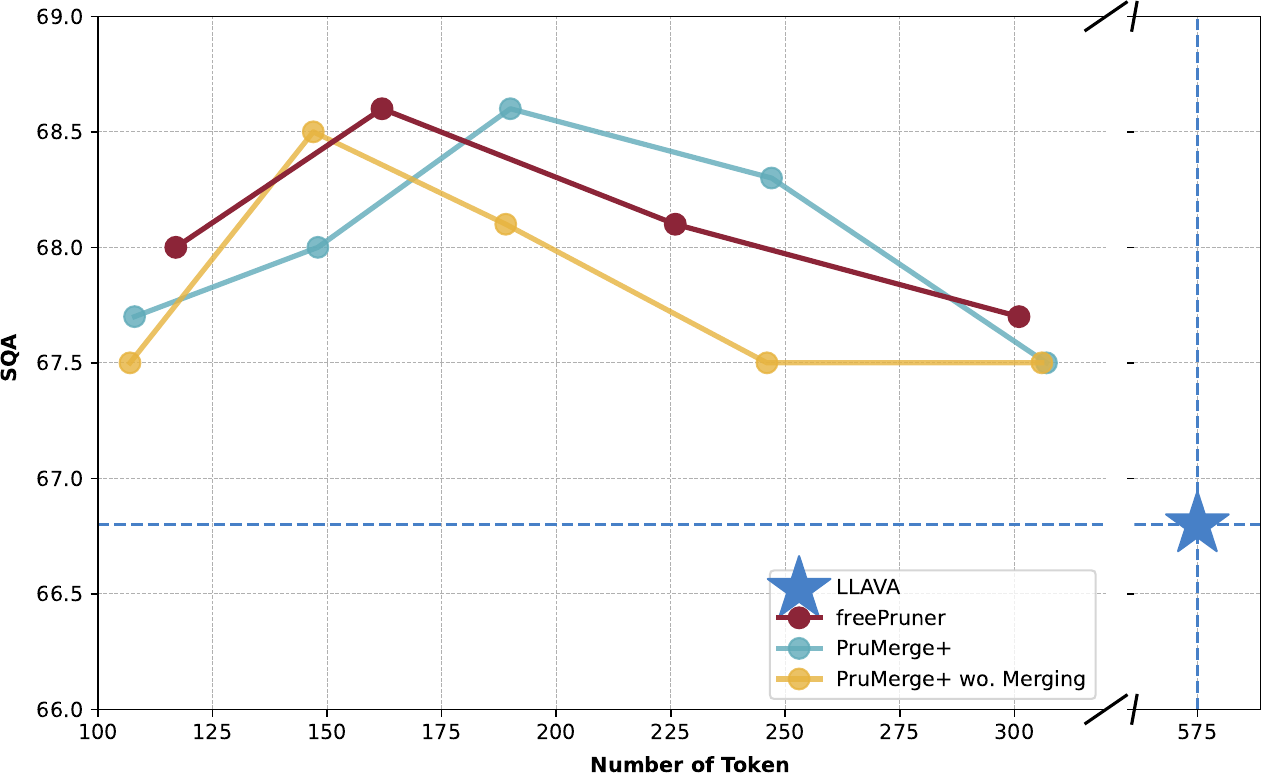}
    \caption{ScienceQA~\citep{lu2022learn}}
    \label{subfig3}
    \end{subfigure}
\end{minipage}
\begin{minipage}{\textwidth}
    \begin{subfigure}{0.33\textwidth}
    \includegraphics[width=\textwidth]{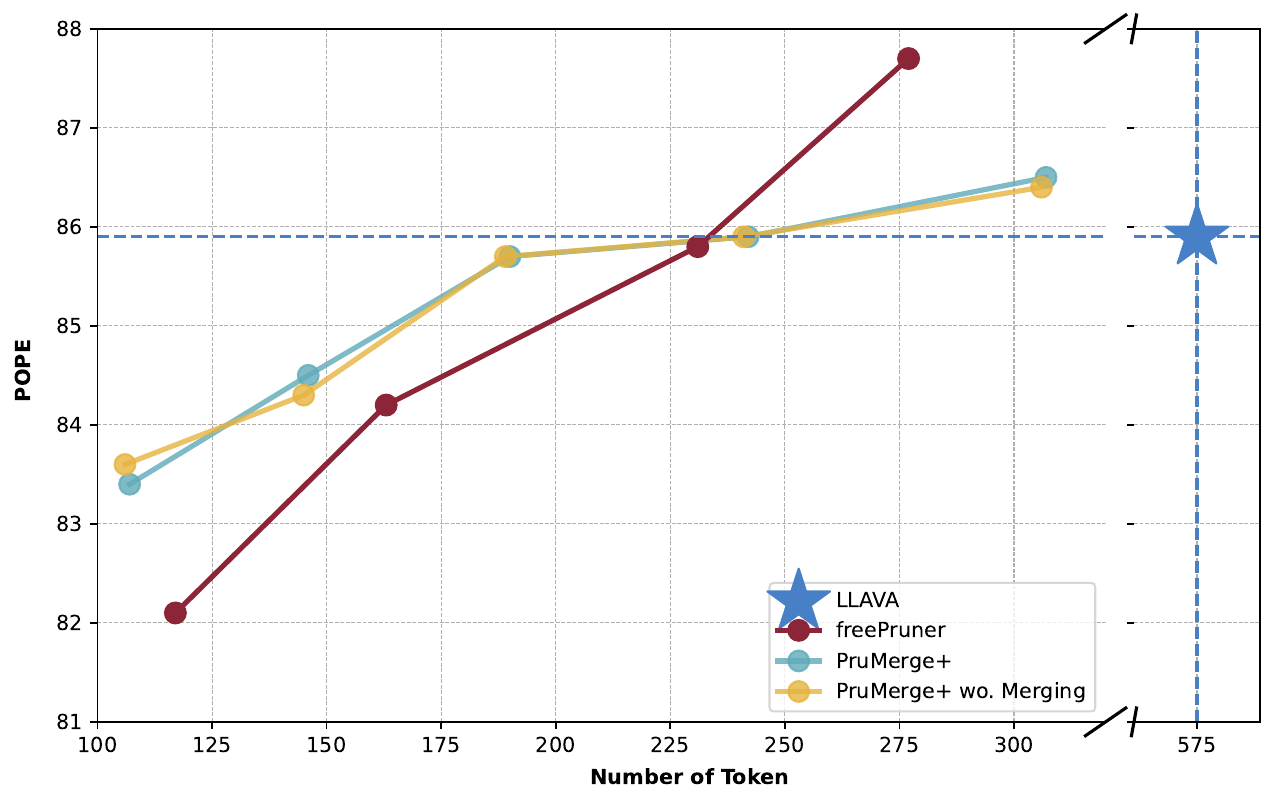}
    \caption{POPE~\citep{li2023pope}}
    \label{subfig4}
    \end{subfigure}
    \hfill
    \begin{subfigure}{0.33\textwidth}
    \includegraphics[width=\textwidth]{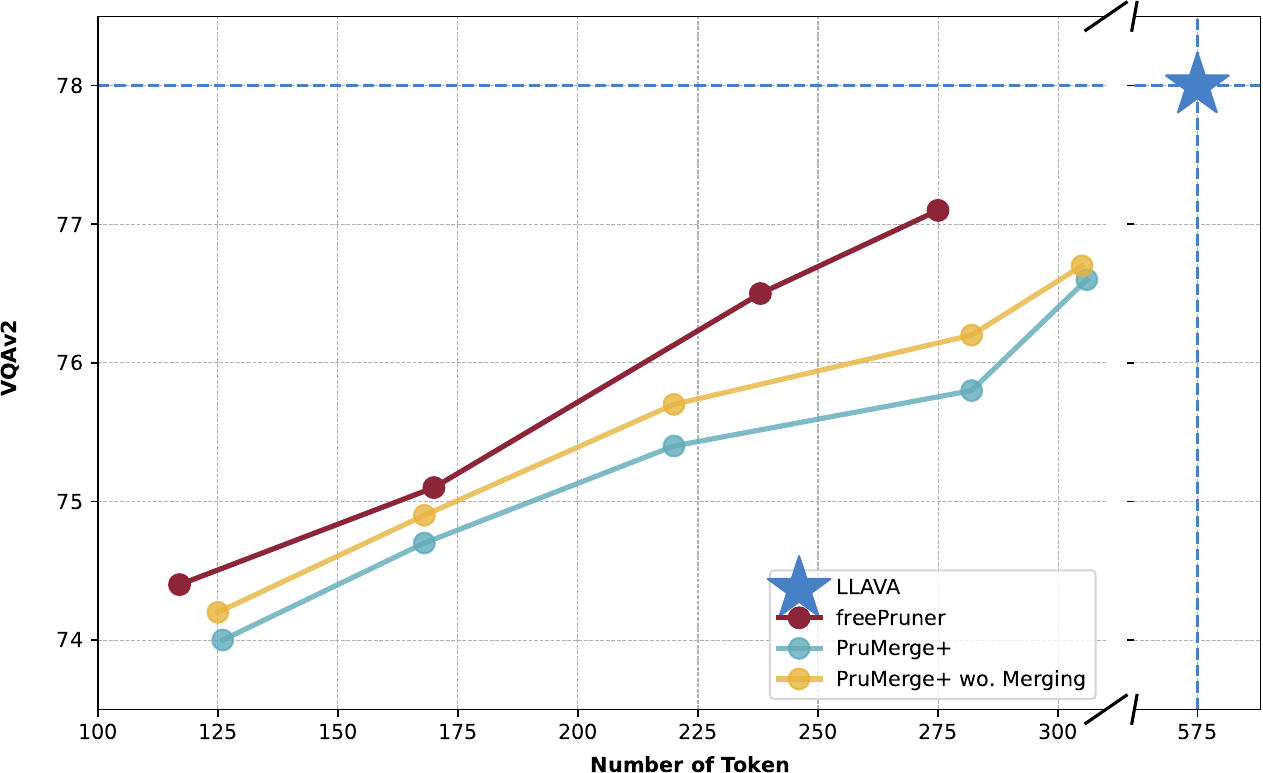}
    \caption{VQAv2~\citep{goyal2017vqav2}}
    \label{subfig5}
    \end{subfigure}
    \hfill
    \begin{subfigure}{0.33\textwidth}
    \includegraphics[width=\textwidth]{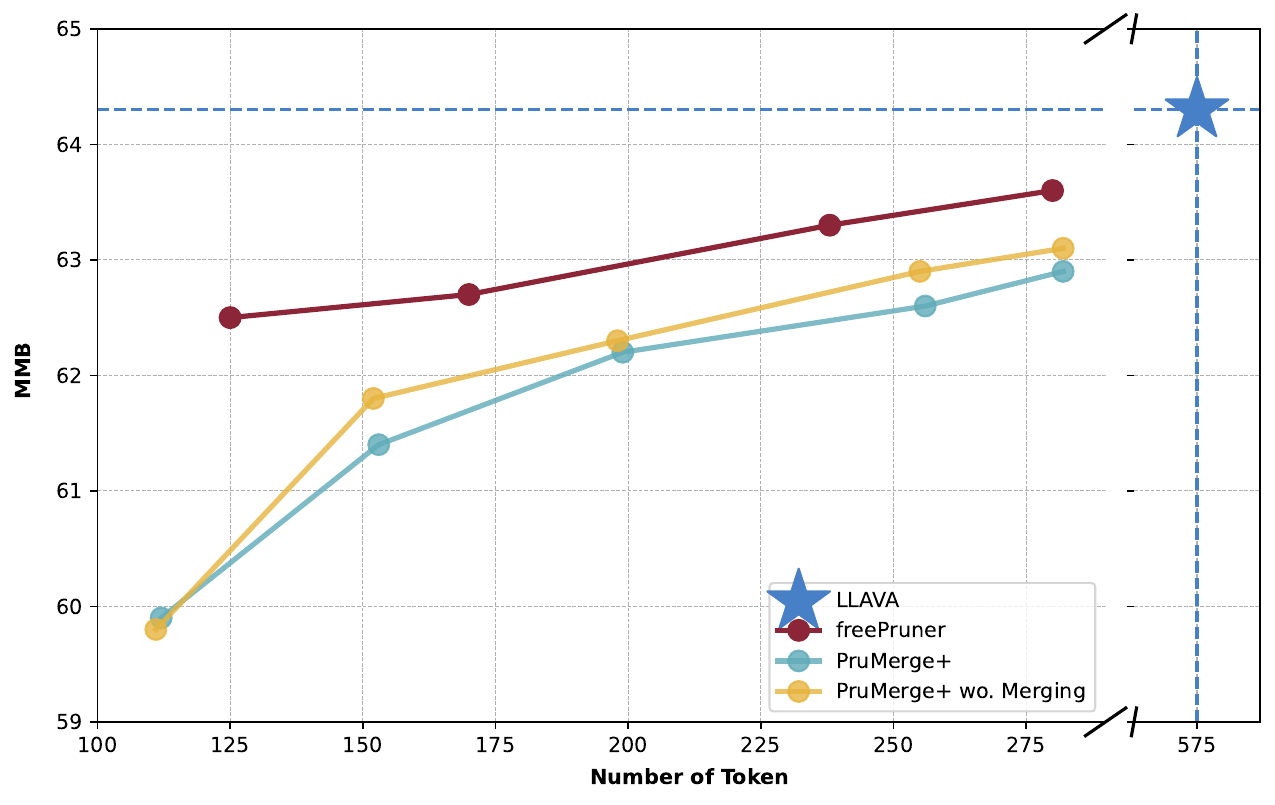}
    \caption{MMBench~\citep{liu2023mmbench}}
    \label{subfig6}
    \end{subfigure}
\end{minipage}
\caption{As the selected number of tokens increase, the performance of \prumergepp~improves, showcasing its scalability. Across various benchmarks, \prumergepp~outperforms \prumerge+~and has comparable performance with LLaVA-1.5.}
\label{fig:result_curve}
\vspace{-0.1in}
\end{figure*} 

\subsection{Main Results}
\label{subsec: Main Results}

We implement our token reduction method to LLaVA-1.5~\citep{liu2023improvedllava}, selecting 50\% of original visual tokens without any training or fine-tuning. Our evaluation spans six visual question-answering and reasoning benchmarks, including VQAv2~\citep{goyal2017vqav2},  ScienceQA~\citep{lu2022learn}, TextVQA~\citep{singh2019textvqa}, POPE hallucination bench~\citep{li2023pope}, MME~\citep{fu2023mme}, and MMBench~\citep{liu2023mmbench}.

As shown in Tab.\ref{tab:main_table}, our approach not only has comparable performance with LLaVA-1.5, but also surpasses it in specific benchmarks like POPE~\citep{li2023pope} and ScienceQA~\citep{lu2022learn} across both 7B and 13B LLM backbones. Comparing with existing token reduction method PruMerge+~\citep{shang2024llava}, \prumergepp~ outperforms its trained and untrained version. Furthermore, our training-free \prumergepp~method outperforms previous models that require training, such as BLIP2~\citep{Li2023BLIP2BL}, InstructBLIP~\citep{instructblip}, Shikra~\citep{chen2023shikra}, and both IDEFICS-9B~\cite{IDEFICS} and IDEFICS-13B~\cite{IDEFICS}. 

\subsection{Scalability Analysis}
\label{subsec: Scalability Analysis}
We explore the scalability of our \prumergepp~method, focusing on how increasing the number of selected tokens influences performance on six VQA benchmarks. As illustrated in Fig.\ref{fig:result_curve}, the model demonstrates improved performance as the number of selected tokens increases on most benchmarks, except for ScienceQA~\citep{lu2022learn}. Additionally, we compare our method against the existing token reduction technique, PruMerge+~\citep{shang2024llava}, using both its merged and non-merged versions as baselines. Our findings reveal that \prumergepp~ consistently outperforms PruMerge+~\citep{shang2024llava} across all benchmarks. As the number of selected tokens increases, \prumergepp~shows a pronounced improvement in performance, while PruMerge+~\citep{shang2024llava} exhibits a more gradual enhancement. This underscores \prumergepp’s robust scalability, even in the absence of further training. When employing only 50\% of the selected tokens, \prumergepp~surpasses the baseline performance of LLAVA-1.5 on TextVQA~\citep{singh2019textvqa}, ScienceQA~\citep{lu2022learn}, and POPE~\citep{li2023pope} tasks. 
We would like to highlight our goal: rather than pursuing extreme token reduction ratios, our method focuses more on achieving training-free acceleration without the need for model retraining—effectively providing ``free-lunch'' speedup for LMM inference. This is why we emphasize on the scalability of training-free token reduction methods.

\subsection{Generalization Across Modalities and Encoders}
\label{subsec:generalization}

To evaluate the broader applicability of \prumergepp, we examine its performance on video understanding tasks using VideoLLaVA~\cite{lin2023video} with the LanguageBind~\cite{zhu2023languagebind} encoder, a transformer-based video-language foundation model. 
Results in Tab.\ref{tab:main_table_video} demonstrate that our method maintains or exceeds the original VideoLLaVA performance while significantly reducing token count (8 times). This effectiveness can be attributed to the inherently higher redundancy in video tokens, which manifests as increased sparsity in transformer attention patterns (see Secs.\ref{subsec:pt} and \ref{subsec:ct}). Given that Video-LLMs typically process substantially more tokens than image-based models~\cite{ryoo2024xgen,liu2024world}, our method's success in this domain suggests particular promise for video understanding applications.
\begin{table}[!t]
\centering
\vspace{-0.1in}
\caption{Results on video reasoning tasks with Video-LLMs. Our method is directly used to Video-LLaVA without training.} 
\scalebox{0.54}{
\begin{tabular}{@{}lc|cc|cc|cc@{}}
\toprule
\multirow{2}{*}{Methods}         & \multirow{2}{*
}{LLM size} & \multicolumn{2}{c|}{\textbf{MSVD-QA}} & \multicolumn{2}{c|}{\textbf{MSRVT-QA}}  & \multicolumn{2}{c}{\textbf{ActivityNet-QA}}  \\ 
                &           & Accuracy    & Score        & Accuracy     & Score         & Accuracy    & Score             \\ \hline
FrozenBiLM      & 1B        & 32.2        & -            & 16.8         & -             & 24.7         & -                 \\
VideoChat       & 7B        & 56.3        & 2.8          & 45.0         & 2.5           & -            & 2.2               \\
LLaMA-Adapter   & 7B        & 54.9        & 3.1          & 43.8         & 2.7           & 34.2         & 2.7               \\
Video-LLaMA     & 7B        & 51.6        & 2.5          & 29.6         & 1.8           & 12.4         & 1.1               \\
Video-ChatGPT   & 7B        & 64.9        & 3.3          & 49.3         & 2.8           & 35.2         & 2.7               \\\hline
Video-LLaVA     & 7B        & 70.7 & 3.9 & 59.2 & 3.5  & 45.3 & 3.3 \\
Video-LLaVA + \prumerge+   & 7B        & 71.1 & 3.9 &  59.3 & 3.6 & 47.7 & 3.4 \\ 
\rowcolor{brown!20} Video-LLaVA + \prumergepp   & 7B       & 71.3 & 3.9 & 59.5 & 3.5 & 48.4 & 3.4  \\\bottomrule
\end{tabular}}
\vspace{-0.2in}
\label{tab:main_table_video}
\end{table}

\subsection{Efficiency Analysis}
\label{subsec: efficiency analysis}

\begin{table*}[t]
\centering
\small
\caption{Computation Cost Analysis based on NVIDIA A6000 GPU, estimated by LLM-Viewer~\citep{yuan2024llm} for theoretical performance.}
\scalebox{0.96}{
\begin{tabular}{l|cc|cccc}
\toprule
\multirow{2}{*}{Method} & LLM      & \multirow{2}{*}{Quantization} & OPs& Prefill      & Accessing & Storing\\
                        & Backbone &                                &  (TB) &  Time (ms)  &  Memory (GB)  &  Activation (GB)\\
\midrule
VideoLLaVA & Vicuna-7B                                   & FP16 & 29.4 & 232.2 & 67.6 & 25.7  \\
\rowcolor{brown!20} VideoLLaVA w/ \prumergepp  & Vicuna-7B & FP16 & 7.3 & 52.6 & 20.6 & 3.3  \\\cline{2-7}
VideoLLaVA & Vicuna-7B & INT4                                   & 29.4 &  102.6 & 16.9 & 6.4  \\
\rowcolor{brown!20} VideoLLaVA w/ \prumergepp  & Vicuna-7B & INT4 &  7.3 & 24.7 & 5.2 & 0.8  \\\hline
LLaVA-1.5 & Vicuna-13B & FP16                                   & 15.9 & 112.7 & 39.2 & 6.1  \\
\rowcolor{brown!20} LLaVA-1.5 w/ \prumergepp  & Vicuna-13B & FP16 &  8.2 & 57.1 & 31.5 & 2.5  \\\cline{2-7}
LLaVA-1.5 & Vicuna-13B & INT4                                    & 15.9 & 53.4 & 9.8 & 1.5  \\
\rowcolor{brown!20} LLaVA-1.5 w/ \prumergepp  & Vicuna-13B & INT4 &  8.2 & 27.4 & 7.9 & 0.6  \\
\bottomrule
\end{tabular}}
\vspace{-0.2in}
\label{tab:efficiency}
\end{table*}

We further look into computational efficiency by applying ~\prumergepp~on LLaVA-13B~\cite{liu2024llava} and VideoLLavA~\cite{lin2023video} with A6000 GPU. The theoretical performance result is estimated by roofline-based LLM-Viewer analysis~\citep{yuan2024llm}.

Tab.\ref{tab:efficiency} shows the efficiency improvement by applying~\prumergepp~on both image and video LMMs. For LLaVA-1.5, \prumergepp~halves the visual tokens, which results in a twofold increase in prefill times. In video-LLM, VideoLLaVA, the impact of \prumergepp~is even more substantial. 
By training-freely reducing the visual tokens to just a quarter of the baseline, \prumergepp~achieves a fourfold increase in prefill times. This considerable reduction in token input also leads to a 70\% decrease in memory access and an eightfold reduction in activation storage in video tasks. 
As the Video-LLMs's deployment bottleneck is in the memory consumption for activation/KV-cache~\cite{shang2024interpolating,fu2024challenges}, our token reduction method shows more potential in the Video-LLMs~\cite{ryoo2024xgen}.
By optimizing the number of visual tokens used, \prumergepp~ensures that models perform more efficiently, maintaining high performance across various benchmarks while optimizing resource use, with particularly notable improvements in video tasks.
It is essential to recognize that the benefits of \prumergepp~are not limited to enhancing efficiency alone. The token reduction technique employed by \prumergepp~can integrate with additional acceleration methods for large multimodal models, such as quantization (see Appendix).

\subsection{Ablation Study on Different Modules}
\label{subsec: Ablation Study}

\begin{figure}[!th]
\centering
  \includegraphics[width=0.48\textwidth]{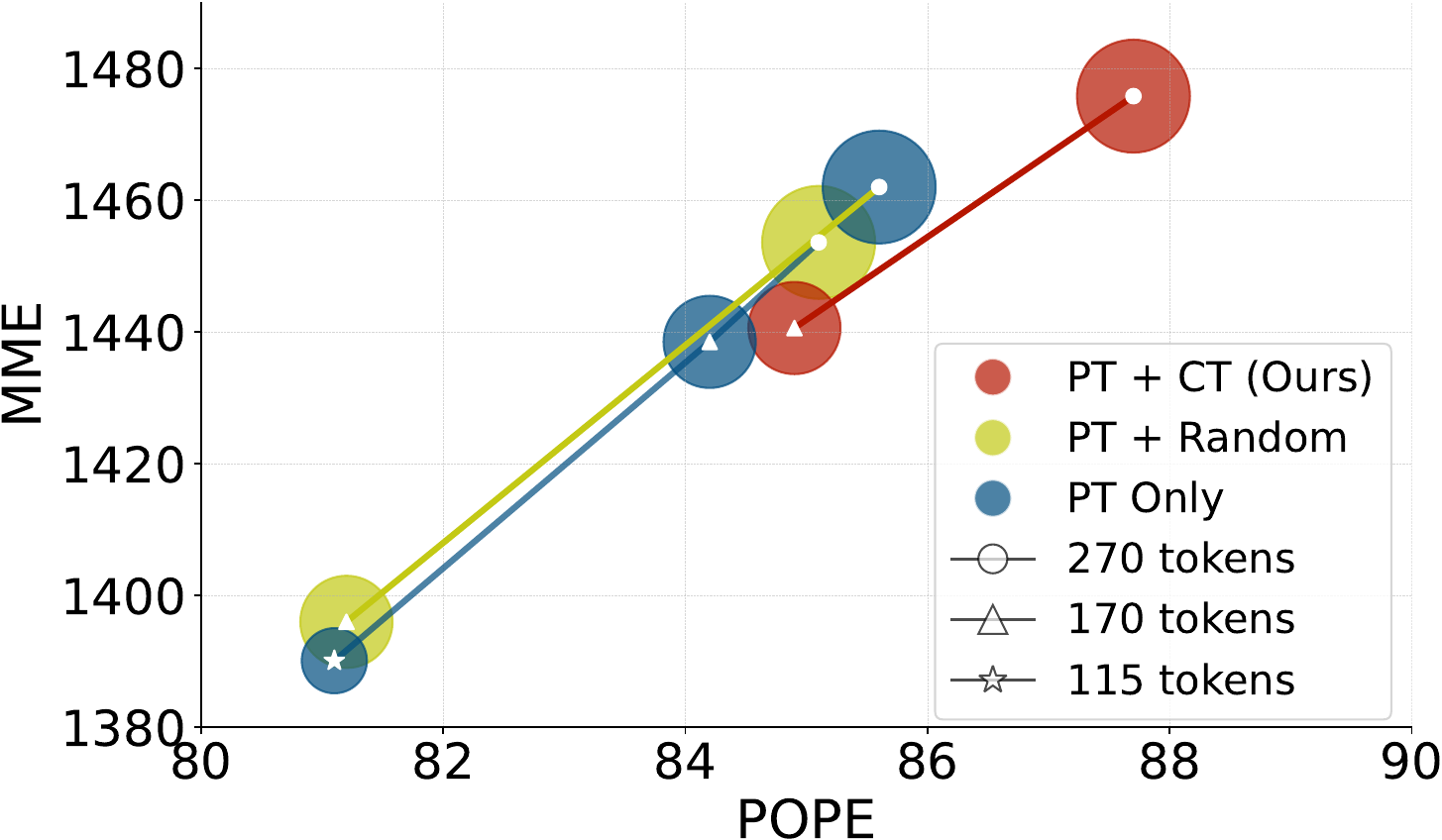}
  \caption{Evaluation of token selection methods on the MME~\cite{fu2023mme} and POPE~\cite{li2023pope} shows that \prumergepp, utilizing both pivotal tokens (PT, Sec.\ref{subsec:pt}) and complementary tokens (CT, Sec.\ref{subsec:ct}), achieves the best performance in scaling the number of tokens. Dot size represents the number of tokens.}
  \vspace{-0.2in}
 \label{fig:ablation_studies_on_ct}
\end{figure}

In this subsection, we examine the effectiveness of the two key modules in \prumergepp~: pivotal tokens (Sec.\ref{subsec:pt}) and complementary tokens (Sec.\ref{subsec:ct}). We aim to validate the necessity of each module and verify whether the coexistence of dual modules is superior to a single module setup. All experiments utilize the LLAVA-1.5 framework with the Vicuna-7B LLM as the backbone.
Fig.\ref{fig:ct_pt} illustrates the performance comparisons among three groups: using only pivotal tokens (PT only), using pivotal tokens plus randomly selected tokens (PT + Random), and using pivotal tokens plus complementary tokens (PT + CT). The pairwise comparison of these groups, shown in Fig.\ref{fig:ablation_studies_on_ct}, reveal three key findings. 
\ul{First, simply increasing the number of tokens does not guarantee a more comprehensive visual representation}. Models using only pivotal tokens outperform those using an equivalent number of tokens composed of pivotal and randomly selected tokens. This suggests that the performance is enhanced by selecting tokens that contain important information. \ul{Second, complementary tokens are not random effects.} We compare two configurations with dual modules, both supported by the same quantity of pivotal tokens but differing in the addition of either complementary tokens or random tokens. The results show that a combination of pivotal and complementary tokens outperforms a mix of pivotal and random tokens. This indicates that the complementary token module provides meaningful information, rather than random effects. \ul{Third, the necessity of complementary tokens is evident; they cannot merely be replaced by increasing the number of pivotal tokens.} Comparisons show that models using both pivotal and complementary tokens outperform those using only pivotal tokens, illustrating that complementary tokens provide crucial low-level visual details that pivotal tokens alone fail to capture.
In summary, both pivotal tokens and complementary tokens presented in our \prumergepp~ method carry important visual information that enhances the overall visual representation.

\vspace{-0.1in}
\section{Conclusion}
\label{sec:con}

We present \prumergepp, a training-free token reduction approach for accelerating LMMs. 
\textit{Rather than pursuing extreme token reduction ratios, our method focuses on achieving training-free acceleration without the need for model retraining—effectively providing ``free-lunch'' speedup for LMM inference.}
We propose a two-stage token selection strategy that identifies and retains the most informative visual tokens while requiring no model retraining or access to training data. Extensive experiments demonstrate that \prumergepp~achieves 2× acceleration while maintaining comparable performance across various visual question-answering tasks. Moreover, our approach is orthogonal to existing acceleration techniques like quantization, offering additional pathways for efficient LMM deployment.

\clearpage


{\small
\bibliographystyle{ieee_fullname}
\bibliography{egbib}
}

\clearpage
\section{Appendix}

\subsection{Orthogonal Method with Post-training Quantization on LLMs}
\begin{table}[!th]
    \centering
    \caption{Results with Quantization Baselines.}
    \scalebox{0.59}{
    \begin{tabular}{cl|ccc|ccc|c}
        \toprule
       \multirow{2}{*}{Bits}  & \multirow{2}{*}{Method} & \multicolumn{3}{c|}{Subject} & \multicolumn{3}{c|}{Context Modality} & Average \\
        &  & NAT & SOC & LAN & TXT & IMG & NO & \\
        \hline
        FP & - & 89.39 & 96.06 & 85.64 & 88.71 & 87.65 & 88.50 & 89.81 \\\hline
        \multirow{3}{*}{W6A6} & AWQ & 85.39 & 92.01 & 83.27 & 84.80 & 83.54 & 85.99 & 86.23 \\
        & QLoRA & 88.45 & 94.71 & 84.48 & 87.63 & 86.07 & 87.87 & 88.43 \\
        & Q-VLM & 89.43 & 95.73 & 84.00 & 88.71 & 87.51 & 87.25 & 89.34 \\
        & Q-VLM + \prumergepp & 89.27 & 95.73 & 84.24 & 88.26 & 86.94 & 87.65 & 88.68 \\
        \hline
        \multirow{3}{*}{W4A4} & AWQ & 74.33 & 72.22 & 74.82 & 73.41 & 67.13 & 77.98 & 74.02 \\
        & QLoRA & 77.53 & 75.48 & 79.18 & 76.64 & 70.70 & 81.95 & 77.53 \\
        & Q-VLM & 80.86 & 75.93 & 80.73 & 80.01 & 72.48 & 83.90 & 79.79 \\
        & Q-VLM + \prumergepp & 81.21 & 75.48 & 80.70 & 81.30 & 72.52 & 83.01 & 79.03 \\
        \bottomrule
    \end{tabular}}
    \label{tab:vlm_quant}
\end{table}

To demonstrate the versatility of \prumergepp, we evaluated its compatibility with existing training-free LLM acceleration techniques, particularly post-training quantization methods. We conducted experiments combining our approach with several prominent quantization methods: AWQ~\cite{lin2024awq}, QLoRA~\cite{dettmers2024qlora}, and the multimodal-specific Q-VLM~\cite{wang2024q}.
Results presented in Table~\ref{tab:vlm_quant} show that \prumergepp~successfully integrates with Q-VLM-quantized LLaVA models~\cite{wang2024q}, functioning as a plug-and-play module without requiring additional modifications. This seamless integration demonstrates the orthogonality of our token reduction approach to existing post-training acceleration methods, suggesting promising opportunities for combining multiple acceleration strategies in the LMM optimization pipeline.

\subsection{\prumergepp~on LLaVA-Next with AnyRes}
\begin{table}[!th]
\centering
\caption{Results of \prumergepp~on LLaVA-Next.} 
\scalebox{0.63}{
\begin{tabular}{lllcc|cc}
\toprule
Method & LLM & Res. & PT & IT & MME & MMB \\
\midrule
LLaVA-Next & Vicuna-7B & 336 & 558K & 665K & 1519.3 & 65.6 \\
LLaVA-Next + \prumergepp & Vicuna-7B & 336 & 0 & 0 & 1502.8 & 64.2  \\
\bottomrule
\end{tabular}}
\label{tab:anyres}
\end{table}

\noindent\textbf{Effectiveness on High-Resolution LMMs.}
LLaVA-Next~\cite{liu2024llavanext} introduced the ``AnyRes'' technique to effectively process high-resolution images while maintaining data efficiency. This capability enables the model to capture fine-grained visual details, significantly reducing hallucination artifacts that typically occur when models process low-resolution inputs. The architecture's ability to handle variable high-resolution inputs makes it particularly valuable for detailed visual analysis tasks.
We evaluated \prumergepp's compatibility with LLaVA-Next's high-resolution processing capabilities, as shown in Table~\ref{tab:anyres}. Note that in our implementation, we modified the standard approach by disabling adaptive token pruning and instead maintaining a fixed token length to align with the AnyRes architecture. Our method successfully reduces the token count by 50\% while preserving the model's high-resolution processing capabilities, effectively doubling the inference speed of LLaVA-Next without compromising its ability to capture fine-grained visual details.

\end{document}